\documentclass[conference]{IEEEtran}
\IEEEoverridecommandlockouts

\usepackage{cite}
\usepackage{amsmath,amssymb,amsfonts}
\usepackage{graphicx}
\usepackage{textcomp}
\usepackage{xcolor}
\def\BibTeX{{\rm B\kern-.05em{\sc i\kern-.025em b}\kern-.08em
    T\kern-.1667em\lower.7ex\hbox{E}\kern-.125emX}}

\usepackage{float}
\usepackage{listings}
\usepackage{csquotes}
\usepackage{multirow}
\usepackage{subcaption}
\usepackage{wrapfig}
\usepackage{arydshln}
\usepackage{booktabs}
\usepackage{hyperref}

\definecolor{darkblue}{rgb}{0, 0, 0.5}
\hypersetup{colorlinks=true, citecolor=darkblue, linkcolor=darkblue, urlcolor=darkblue}
\usepackage{microtype}
\usepackage{url}
\usepackage{lineno}
\usepackage{algorithm}
\usepackage{algpseudocode}

\begin{document}
\bstctlcite{IEEEexample:BSTcontrol}

\IEEEpubid{\makebox[\columnwidth]{IEEE ICKG 2025. \copyright\ 2025 IEEE \hfill}
\hspace{\columnsep}\makebox[\columnwidth]{ }}

\title{Can an LLM Induce a Graph?\\Investigating Memory Drift and Context Length}

\author{\IEEEauthorblockN{Raquib Bin Yousuf, Aadyant Khatri, Shengzhe Xu, Mandar Sharma, Naren Ramakrishnan} \IEEEauthorblockA{Department of Computer Science, Virginia Tech, Alexandria, VA\\
Email: raquib@vt.edu, naren@cs.vt.edu}}

\maketitle

\begin{abstract}
Recently proposed evaluation benchmarks aim to characterize the effective context length and the forgetting tendencies of large language models (LLMs). However, these benchmarks often rely on simplistic ``needle in a haystack'' retrieval or continuation tasks that may not accurately reflect the performance of these models in information-dense scenarios. Thus, rather than simple next token prediction, we argue for evaluating these models on more complex reasoning tasks that requires them to induce structured relational knowledge from the text - such as graphs from potentially noisy natural language content. While the input text can be viewed as generated in terms of a graph, its structure is not made explicit and connections must be induced from distributed textual cues, separated by long contexts and interspersed with irrelevant information. Our findings reveal that LLMs begin to exhibit memory drift and contextual forgetting at much shorter effective lengths when tasked with this form of relational reasoning, compared to what existing benchmarks suggest. With these findings, we offer recommendations for the optimal use of popular LLMs for complex reasoning tasks. We further show that even models specialized for reasoning, such as OpenAI o1, remain vulnerable to early memory drift in these settings. These results point to significant limitations in the models' ability to abstract structured knowledge from unstructured input and highlight the need for architectural adaptations to improve long-range reasoning.
Our codebase to support reproducibility is publicly available.\footnote{\url{https://github.com/DiscoveryAnalyticsCenter/MemoryDrift}}.
\end{abstract}

\begin{IEEEkeywords}
benchmark, evaluation, contextual forgetting, context length, memory drift
\end{IEEEkeywords}

\section{Introduction}
\label{sec:intro}
Recent benchmarks for evaluating LLMs have made significant progress in measuring context length and memory retention \cite{ karpinska2024one, levy2024same, song2024counting}. However, many of these evaluations rely on highly synthetic tasks, such as ``needle-in-a-haystack'' retrieval \cite{hsieh2024ruler, needleinahaystack_2024, li2024needlebench} or shallow continuation \cite{liu2024forgetting}, which do not reflect the kinds of structured reasoning and information integration required in practical applications. While several of these works acknowledge the limitations of such tasks, particularly in capturing realistic comprehension or reasoning demands \cite{hsieh2024ruler, li2024needlebench, liu2024forgetting}, they still fall short of evaluating whether a model can \textit{induce latent structure} from long and noisy text.
\begin{figure}[t]
    \centering
    \includegraphics[width=\columnwidth]{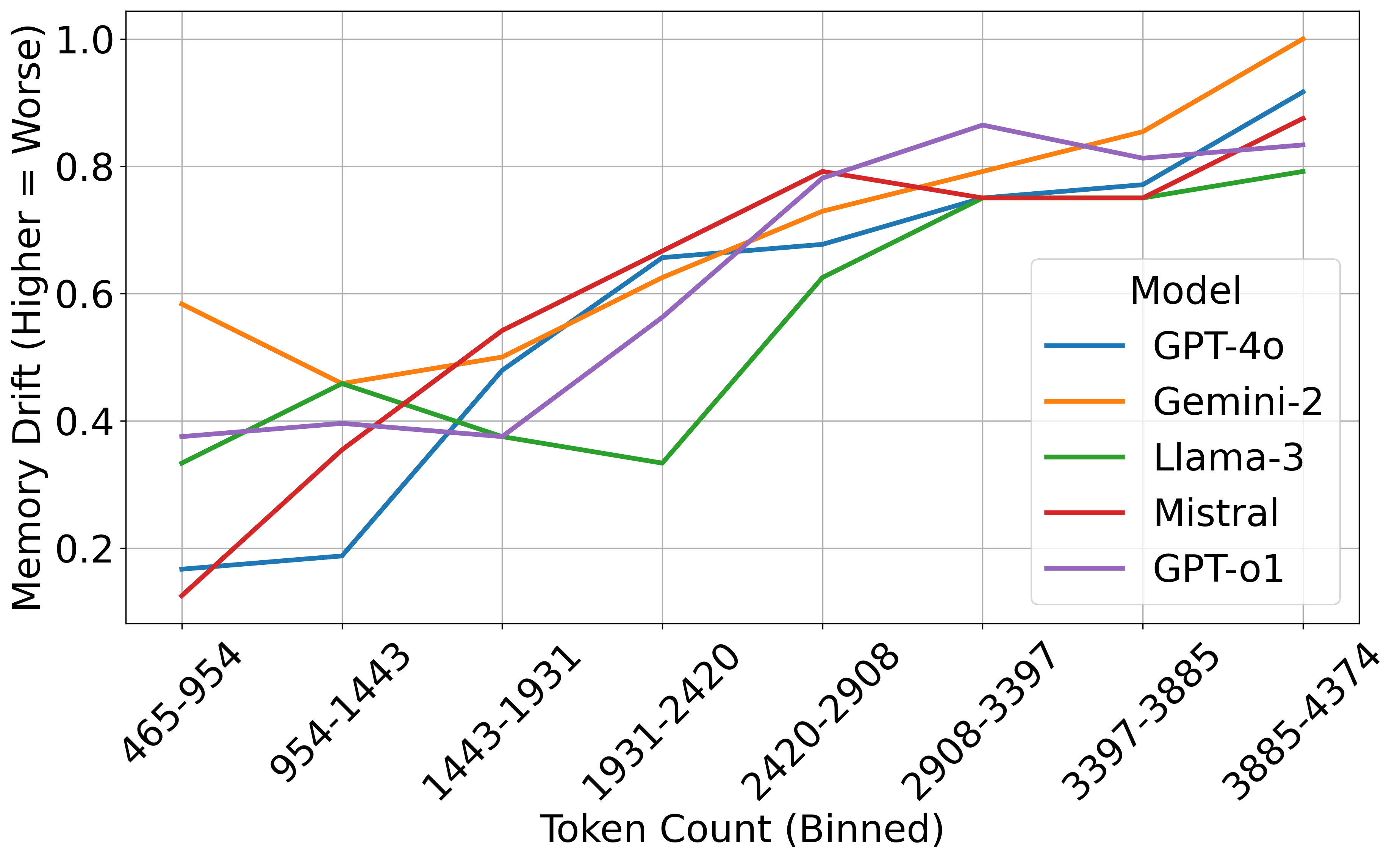}
    \caption{Memory drift (lower = better) on the simplest relational task (one connection per sample). Despite minimal complexity, all models degrade beyond a certain context length, showing that even low relational load challenges long-context reasoning. See Table~\ref{tab:model_comparison} for TL;DR.}
    \label{fig:entry_overlay}
\end{figure}
\begin{figure*}[t]
    \centering
\includegraphics[width=\textwidth]{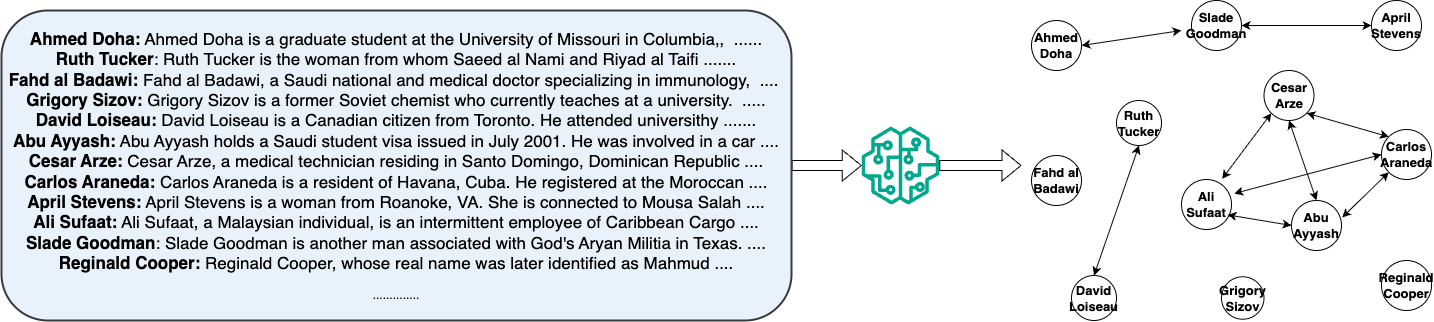}
    \caption{Overview of our task: given a long, noisy text (left), the model (center) reconstructs the underlying relational graph (right) by identifying connections between entities (edge, subgraph, clique). Disconnected nodes are distractors.}
    \label{fig:entry_task}
\end{figure*}

In contrast, real-world reasoning often requires connecting entities and events that are scattered across large, unstructured documents \cite{shankar2024docetl, yousuf2024llm, fan2024medodyssey}. The relevant relationships are rarely local or explicit, but must be inferred from distributed and indirect cues. Whether in scientific literature review, legal understanding, 
intelligence analysis, or medical report comprehension, effective reasoning involves recovering sparse relational knowledge embedded within irrelevant content. Evaluating models only on token-level recall or completion fails to capture this challenge.

In this work, we argue that the effective context length and the forgetting tendencies of LLMs should be evaluated based on their ability to recover \textit{relational graphs} from natural language. These graphs encode semantic connections between entities, events, or concepts, and serve as a cognitively aligned abstraction of real-world information needs. Crucially, the graph structure is not provided directly, and must be inferred from paraphrased, interleaved, and noisy textual descriptions.

To this end, we introduce a new benchmark for evaluating the effective context length and the forgetting tendencies of LLMs centered on \textbf{graph reconstruction from noisy text}. Given a long input that implicitly encodes a hidden graph, the model must identify the correct nodes and their pairwise relations. We systematically control two axes of difficulty: (i) \textit{contextual separation}, which measures how far apart related entities appear in the prompt, and (ii) \textit{relational density}, which quantifies the number of connections the model must recover. These controls allow us to probe how models degrade under increased memory stress and structural complexity.

Our empirical findings reveal several consistent patterns in how large language models handle long-context relational reasoning:

\begin{itemize}
    \item \textbf{Onset of memory drift at shorter effective context lengths.} Across models, performance degradation begins well before the maximum supported context window, with measurable declines in structural recovery observed as the length and complexity of input increases.

    \item \textbf{Recall is often the limiting factor in relational reasoning.} Many models favor high-precision extraction strategies, resulting in a tendency to miss valid connections rather than produce spurious ones. This conservative behavior becomes more pronounced as context length grows.

    \item \textbf{Greater structural complexity amplifies degradation.} As the density of relationships or the number of connections within each prompt increases, all models display declining performance, indicating heightened sensitivity to relational complexity in long-context scenarios.

    \item \textbf{Prompting strategies such as chain-of-thought do not mitigate these challenges.} Experiments with different prompting styles, including chain-of-thought reasoning, show little or no improvement for long-context graph reconstruction tasks and can, in some cases, worsen results due to increased distraction.

    \item \textbf{A reasoning-specialized model does not overcome early memory drift.} In our evaluation, a model specifically designed for advanced reasoning also exhibited early onset of memory drift, similar to general-purpose language models. This suggests that, at least for the tested model, advanced reasoning capabilities alone are insufficient to address the core limitations of long-context relational reasoning.
\end{itemize}

These results highlight persistent brittleness in current LLMs when faced with long-context relational reasoning, particularly when structural recovery is required under dispersion, density, and noise. Recovering graph structure from noisy text is substantially more demanding than next-token prediction or span-level retrieval. These observations underscore the need to move beyond generic context-length benchmarks toward task-specific evaluations that reflect structured reasoning demands. As summarized in Table~\ref{tab:model_comparison}, models vary significantly in how they balance precision, recall, and memory stability, which emphasizes the importance of targeted model selection for real-world applications. Our key contributions are:

\begin{itemize}
    \item We design a graph reconstruction tasks for LLMs, consisting three subtasks, edge recovery, subgraph discovery, and clique detection, that probe a model’s ability to induce structure under dispersion and noise.
    
    \item We propose \textit{memory drift}, a metric that captures forgetting and hallucination as a function of context length and relational complexity.

    \item We systematically evaluate five popular LLMs (GPT-4o \cite{hurst2024gpt-4o}, OpenAI o1 \cite{o1_intro, jaech2024openai_o1}, Gemini-2 \cite{team2024gemini}, Llama-3 \cite{grattafiori2024llama}, Mistral-7B \cite{jiang2023mistral7b}), revealing earlier and sharper degradation than suggested by existing benchmarks, especially under high information density.

    \item We release our codebase to support reproducible analysis of long-context reasoning via structured knowledge extraction.
\end{itemize}

\section{Background and Related Work}
\subsection{Context Length and Memory in Language Models}
\label{sec:related_work_memory}
Recent years have seen rapid progress in the evaluation of large language models (LLMs) on long-context understanding. Several benchmarks have sought to quantify the effective memory and forgetting behavior of LLMs using controlled experimental designs. For instance, the Forgetting Curve~\cite{liu2024forgetting}, Same Task, More Token~\cite{levy2024same}, One Thousand and One Pairs~\cite{karpinska2024one}, and Ruler~\cite{hsieh2024ruler} benchmarks probe the extent to which LLMs can retrieve information or maintain associations over increasing input lengths.

However, these efforts primarily use synthetic or simplified tasks, and often do not reflect model performance in settings where relevant information is sparsely distributed or interleaved with distractors. The true extent of early memory drift and contextual forgetting in more complex, relational reasoning tasks remains underexplored, motivating our present study.

\subsection{Relational Reasoning with LLMs}
LLMs have progressed from basic text generation~\cite{achiam2023gpt, touvron2023llama} to complex applications involving chat agents~\cite{park2023generative}, multi-agent simulation~\cite{wang2023voyager}, and scientific reasoning~\cite{romera2024mathematical}. Structured information extraction and relational reasoning have become increasingly important, particularly in domains such as intelligence analysis~\cite{yousuf2024llm, tang2024steering}, where models must recover key relationships embedded in lengthy, noisy text. 

A core challenge in these applications is identifying salient clues and mapping entity relationships across large, unstructured inputs~\cite{lu2024chameleon_tool, schick2024toolformer, zhang2023graphtoolformer}. Recent approaches leverage tool augmentation or retrieval~\cite{lewis2020retrieval, zhong2022training, wang2023selfknowledge, gao2023retrieval, shuster2022language} to supplement the model’s latent knowledge. Other work explores the capacity of LLMs to maintain organized memory structures with or without external augmentation~\cite{yousuf2024llm}. 

Despite these advances, existing evidence suggests that even advanced LLMs struggle with long-context relational reasoning, especially when structural cues are dispersed or implicit. In intelligence analysis tasks, for example, effective memory length may be even shorter than in conventional text benchmarks~\cite{liu2024forgetting, levy2024same, karpinska2024one, hsieh2024ruler}, and the ability to induce latent graph structure remains a significant limitation.

\subsection{Relation to Entity Linking, Coreference Resolution, and Knowledge Base Construction}
While our benchmark is superficially related to traditional NLP tasks such as entity linking, coreference resolution, relation extraction, and knowledge base construction (KBC), there are fundamental differences in both objective and methodological focus. In recent years, large language models (LLMs) have been increasingly applied to these classical tasks~\cite{xin2024llmael, onenet2024, lelanguage, oshima2024synthetic, liu2024bridging, min2024synergetic, sundar2024major, li2024relation, tao2024graphical, hu2024gptkb, nayak2023llm2kb, singhania2022lmkbc}. Entity linking, coreference resolution (mention clustering/anaphora), and relation extraction have all seen improvements from generative modeling, instruction tuning, and prompt-based LLMs. Some recent studies have further explored the limitations of LLMs with respect to context length and memory for these tasks, particularly as applied to longer documents or document-level extraction~\cite{zhu2025llmlink, liu2024bridging, nayak2023llm2kb, li2024long}.

However, the majority of this prior work continues to focus on local mention disambiguation, anaphora resolution, or extraction of predefined relation types, typically in settings where relevant cues are assumed to co-occur or be easily retrievable within limited context windows. In contrast, our benchmark diverges in both its central aim and experimental design. We treat \textbf{relational graph reconstruction as a direct proxy for analyzing LLM memory, context length, and forgetting in information-dense, noisy settings}. Specifically, our task requires models to process extended and noisy input sequences, where relational cues are highly dispersed, indirect, and embedded within substantial irrelevant content. The model must encode and maintain distributed entity descriptions over long-range dependencies, and integrate these representations to induce latent connections, often separated by significant contextual distance or paraphrased evidence. This setting compels holistic, graph-level reasoning rather than isolated or span-local predictions.

Crucially, our benchmark is structured to push models to their effective memory and reasoning limits by (i) dispersing relational evidence across extended contexts, (ii) interleaving structurally irrelevant distractors, and (iii) increasing the density and granularity of latent relational structure that must be reconstructed jointly. These design choices go beyond the scope of existing information extraction or KBC benchmarks and provide a more rigorous test of long-context reasoning.

Taken together, while recent advances have extended the reach of LLMs in structured information extraction and relational reasoning, existing benchmarks do not adequately capture the challenges posed by long, noisy contexts where latent structure must be recovered globally. To fill this gap, we introduce \textbf{relational graph reconstruction} as a general probe of long-context reasoning and memory in LLMs, enabling systematic evaluation of their ability to integrate dispersed, indirect, and noisy relational cues across extended input sequences.

\section{Graph Reconstruction as a Lens on Long-Context Reasoning}
\label{sec:method}
Despite recent progress in information extraction and reasoning, it remains unclear whether LLMs can integrate and recover latent relational structure from long, noisy, and unstructured inputs. We address this by treating graph reconstruction as a direct probe of long-context memory and reasoning in LLMs, using data inspired by real-world, intelligence-style reporting.

The core task of our benchmark is \textbf{relational graph reconstruction}. Given a long natural language input encoding a hidden graph, the model must recover this graph through structured prediction or post-hoc extraction. The key challenge lies in piecing together distributed cues that correspond to nodes and their connections. Natural language rarely encodes explicit graph edges. Instead, relations are embedded in distributed mentions, paraphrases, and disjoint spans. Moreover, real-world text contains distractors or irrelevant facts, entities, or events, which the model must learn to ignore. The presence of such noise compounds the difficulty of maintaining stable memory traces over long sequences.

Thus, with the above consideration, we evaluate three subtasks: \textbf{(i) Edge Discovery}, where the model recovers pairwise relations; \textbf{(ii) Subgraph Discovery}, where it identifies connected node subsets (e.g., stars, chains); and \textbf{(iii) Clique Discovery}, where it detects fully connected clusters.

\begin{figure*}[!ht]
    \centering
    \includegraphics[width=\textwidth]{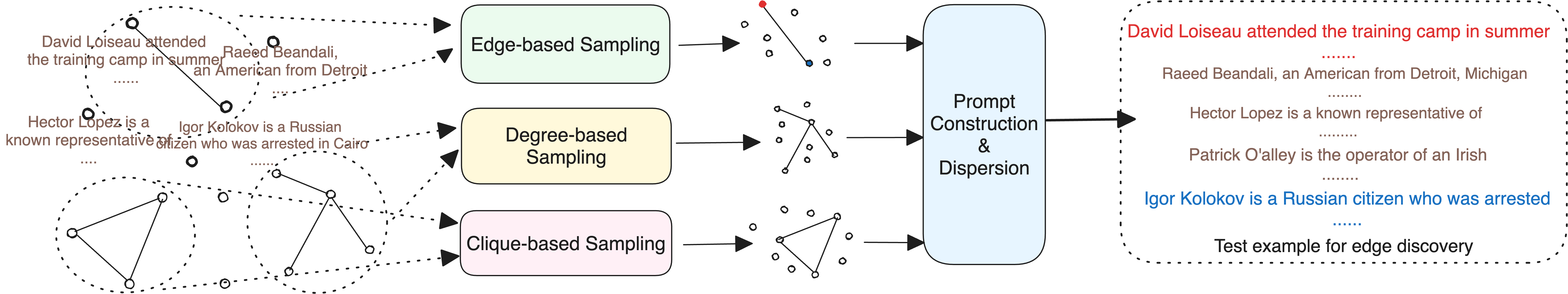}
    \caption{Graph-to-prompt pipeline: relational structures (edges, stars, cliques) are sampled from a latent graph and interleaved with distractors to create dispersed, noisy prompts. Red and Blue are targets. Brown are the distractors.}
    \label{fig:graph_to_prompt}
\end{figure*}

\begin{table*}[ht]
\caption{Summary of graph-based tasks used in the benchmark. Each task varies in structural complexity and connection granularity.}
\centering
\resizebox{0.90\textwidth}{!}{%
\begin{tabular}{@{}llll@{}}
\toprule
\textbf{Task Type}        & \textbf{Structure}           & \textbf{Connection Size}              & \textbf{Sampling Strategy} \\
\midrule
Edge Discovery            & Pairwise edges $(u, v)$      & 2 nodes per connection                & Min-degree edge selection \\
Subgraph Discovery        & Star-like subgraphs          & $d + 1$ nodes (degree-based)          & Min-degree neighborhood centered on node \\
Clique Discovery          & Fully connected cliques      & $k$ nodes per clique                  & Min-degree clique removal \\
\bottomrule
\end{tabular}
}
\label{tab:task-summary}
\end{table*}

\subsection{Task Formulation}
Let $\mathcal{G} = (V, E)$ be an undirected latent graph over a set of entities $V$, where each edge $(u, v) \in E$ represents a hidden semantic relation. Each node $v \in V$ has a corresponding natural language description $d_v \in \mathcal{D}$.

\textbf{Input Construction:} We define a prompt $\Pi = \{d_{v_1}, \dots, d_{v_n}\}$ where $n = |\mathcal{C}| + |\mathcal{N}|$, composed of:
\begin{itemize}
    \item $\mathcal{C}$: a set of connected components drawn from $\mathcal{G}$ (e.g., edges, stars, cliques),
    \item $\mathcal{N}$: a set of noise or distractor nodes such that $\forall u,v \in \mathcal{N}$, $(u,v) \notin E$.
\end{itemize}

Let $\texttt{disp}: \mathcal{C} \times \mathcal{N} \rightarrow \Pi$ be a dispersion function that interleaves elements of $\mathcal{C}$ into the distractor set, controlling their relative positions.

We define the token-level separation between two related entities $u,v \in \mathcal{C}$ in the prompt $\Pi$ as:
\[
\delta(u, v; \Pi) = \texttt{TokenDist}(d_u, d_v)
\]
and use this to measure contextual dispersion or memory stress.

\textbf{Language Model Objective:} Let $f_\theta: \Pi \rightarrow \hat{\mathcal{G}} = (\hat{V}, \hat{E})$ be the output of the language model, where the goal is to recover:
\[
\hat{\mathcal{G}} \approx \mathcal{G}[\mathcal{C}]
\]
i.e., reconstruct the subgraph induced by the true connected components.

\textbf{Evaluation:} We evaluate predictions using a graph-level reconstruction loss:
\[
\mathcal{L}(\hat{\mathcal{G}}, \mathcal{G}[\mathcal{C}]) = \texttt{F1}_{\text{edge}}(\hat{E}, E[\mathcal{C}])
\]
alongside \textit{memory drift} metrics based on $\delta(u,v)$ and performance degradation across increasing distance.

\textbf{Benchmark Variants:} We instantiate this task under three structural regimes:
\begin{itemize}
    \item \textbf{Edge Discovery:} $\mathcal{C} = \{(u_i, v_i)\}_{i=1}^k$ where each $(u_i, v_i) \in E$.
    \item \textbf{Subgraph Discovery:} $\mathcal{C}$ contains $k$ star-like subgraphs of size $d+1$.
    \item \textbf{Clique Discovery:} $\mathcal{C}$ contains $k$ cliques of size $k$ such that all $(u, v) \in \mathcal{C}$ satisfy $(u,v) \in E$.
\end{itemize}

\subsection{Graph Sampling Strategies}
\label{ssec:graph_sampling}
\subsubsection{Edge-Based Sampling (Edge Discovery)}
For this case, we isolate individual pairwise relations for evaluation. Each test instance is constructed by identifying edges that lie near the periphery of the graph, where nodes are less embedded within dense neighborhoods. Formally, we define the priority of an edge $(u, v)$ by the combined degree of its endpoints: $\deg(u, v) = \deg(u) + \deg(v)$. Edges with the smallest such values are selected first, under the assumption that they are less likely to overlap with other relational structures. To prevent redundancy or relational leakage, both the selected edge and its immediate neighbors are excluded from further sampling. This enforces disjointness and ensures that connections are spatially isolated within the graph topology.

Remaining nodes that are no longer part of any edge structure are repurposed as distractors. These nodes do not participate in relational content relevant to the task, and serve as negative examples. This setting provides a base case for evaluating whether models can recover simple binary relations when embedded in distractive and unstructured input.

\subsubsection{Degree-Based Sampling (Subgraph Discovery)}
To examine a model’s ability to recover higher-order structure, we sample local neighborhoods centered around nodes of fixed degree. These subgraphs are star-like, consisting of a central node and its $d$ immediate neighbors, forming an induced subgraph of size $d + 1$. Each candidate neighborhood is scored based on the total degree of its closed neighborhood, i.e., $\sum_{u \in N[v]} \deg(u)$; where $N[v]$ is the set containing $v$ and all nodes adjacent to it. Lower-scoring neighborhoods are prioritized, under the hypothesis that they are less entangled within the larger graph and more likely to be topologically separable.

Once a subgraph is selected, its constituent nodes and all adjacent nodes are removed to maintain disjointness between samples. The remaining portion of the graph is explicitly disconnected, eliminating residual connectivity and yielding a pool of structurally neutral distractors. This task setup probes whether a model can identify subgraphs with coherent internal structure when they are distributed among unrelated textual descriptions.

\subsubsection{Clique-Based Sampling (Clique Discovery)}
The most structurally demanding variant of our benchmark targets the recovery of cliques: fully connected subgraphs where each node shares an edge with every other node in the group. A clique of size $k$ satisfies $(u, v) \in E$ for all $u, v \in C$, where $C = \{v_1, \dots, v_k\}$. Such substructures require the model to integrate multiple overlapping relations simultaneously.

We restrict our attention to cliques that are not only maximal but also situated in sparsely connected regions. Each candidate clique is scored by the aggregate degree of its nodes, i.e., $\sum_{i=1}^{k} \deg(v_i)$. Those with lower scores are preferred, as they are more likely to be separable from the rest of the graph. After sampling, the clique and its surrounding neighborhood are removed to prevent structural overlap between samples.

Distractors are drawn from the remaining graph, which are forcibly pruned to remove all residual edges. The result is a controlled environment in which the only coherent structure is the clique itself. This setting evaluates the model’s ability to recognize dense and mutually entangled relational clusters within noisy, otherwise unstructured contexts. Algorithm \ref{alg:general-sampling} shows a generalized version of three different sampling techniques.

\begin{figure}[h]
\begin{minipage}{\columnwidth}
    \begin{algorithm}[H]
    \small
    \caption{General Subgraph-Based Sampling}
    \label{alg:general-sampling}
    \begin{algorithmic}[1]
    \Require Graph $G = (V, E)$; selector type $\mathsf{Type}$ (e.g., \textsc{Edge}, \textsc{Clique}, \textsc{Degree}); parameter $p$ (e.g., $k$ or $d$)
    \Ensure $\mathcal{C}$: selected subgraphs, $\mathcal{D}$: disconnected nodes
    
    \State $\mathcal{C} \gets [\,]$; $\mathcal{D} \gets [\,]$; $G' \gets G$
    \While{there are valid units of type $\mathsf{Type}$ in $G'$}
        \If{$\mathsf{Type}$ is \textsc{Edge}}
            \State Select $(u,v) \gets \arg\min_{(i,j) \in E(G')} \deg(i) + \deg(j)$
            \State $U \gets \{u,v\}$
        \ElsIf{$\mathsf{Type}$ is \textsc{Clique}}
            \State Find all cliques $\mathcal{Q} = \{C \subseteq V(G') \mid |C|=p, C\ \text{is clique}\}$
            \State $U \gets \arg\min_{C \in \mathcal{Q}} \sum_{v \in C} \deg(v)$
        \ElsIf{$\mathsf{Type}$ is \textsc{Degree}}
            \State $N_p \gets \{v \in V(G') \mid \deg(v) = p\}$
            \State $v^* \gets \arg\min_{v \in N_p} \sum_{u \in N[v]} \deg(u)$
            \State $U \gets N[v^*]$
        \EndIf
        \State $\mathcal{C} \gets \mathcal{C} \cup \{U\}$
        \State $R \gets N[U]$ \Comment{closed neighborhood of $U$}
        \State Remove $R$ from $G'$
        \State $\mathcal{D} \gets \mathcal{D} \cup V(G')$
    \EndWhile
    \State \Return $\mathcal{C}, \mathcal{D}$
    \end{algorithmic}
    \end{algorithm}
\end{minipage}

\begin{minipage}{\columnwidth}
    \begin{algorithm}[H]
    \small
    \caption{Prompt Test Case Generation with Dispersion}
    \label{alg:test-dispersion}
    \begin{algorithmic}[1]
    \Require Profiles $\mathcal{P}$, connection dict $\mathcal{C}$, params $\Theta = \{k, n, s, e\}$, tokenizer $\mathcal{T}$
    \Ensure Test cases $\mathcal{T}_{\text{cases}}$
    \State $\mathcal{T}_{\text{cases}} \gets [\,]$
    \For{$i = 1$ to $N$}
        \State Sample $C = \{c_1, \dots, c_k\}$ from $\mathcal{C}$
        \State Sample $D \subset V$, $|C| + |D| = n$
        \State Partition $D$ into $k$ segments in $[s \cdot |D|, e \cdot |D|]$
        \State Interleave each $c_j$ into segment $j$ of $D$ to form $L$
        \State $\Pi \gets \texttt{Concat}(\mathcal{P}[x]\,|\,x \in L)$
        \State Compute $\delta = \texttt{token\_dist}(c_1, c_k, \mathcal{T})$
        \State Store $\{L, \Pi, \delta\}$ in $\mathcal{T}_{\text{cases}}$
    \EndFor
    \State \Return $\mathcal{T}_{\text{cases}}$
    \end{algorithmic}
    \end{algorithm}
\end{minipage}%
\end{figure}

\subsection{Prompt Construction}
To systematically evaluate long-context relational reasoning, we construct a controlled benchmark that synthesizes input prompts containing both relational and distractor entities. Each test case is generated through a three-stage pipeline: (i) graph sampling to define ground-truth connections, (ii) controlled instantiation of test prompts with a mix of connected and disconnected entities, and (iii) spatial dispersion of related components to simulate contextual separation. 

Given a curated entity graph and corresponding textual profiles, we generate a set of test instances tailored to each task type—edge discovery, subgraph recovery, or clique detection. The prompt generation process samples structured relational subsets and interleaves them with distractor nodes, allowing us to modulate both structural complexity and memory stress.

For edge discovery, each test case includes a fixed number of connected pairs sampled from the graph's edge set. Distractors are drawn from two sources: (a) explicitly disconnected nodes, and (b) unused nodes from the relational pool, with at most one node per unused pair. This guarantees a consistent number of entities per prompt while preserving topological separation between connected and distractor elements.

For subgraph and clique discovery tasks, we first sample tuples of higher cardinality. These are drawn from the graph according to degree-based or clique-specific criteria, ensuring that each tuple forms a valid induced substructure. The number of nodes per substructure is varied within a defined range, and samples are stratified by structural regime (e.g., stars or cliques). Each test case includes a balanced mixture of such substructures and distractors, with a fixed total entity budget.

Finally, for every sampled prompt, we compute the adjacency matrix of the ground-truth subgraph to enable evaluation. The relative placement of connected entities within the distractor pool is explicitly controlled to vary contextual separation and dispersion, allowing us to probe the model’s ability to integrate non-local relational cues. Algorithm \ref{alg:test-dispersion} shows the inner-works of prompt generation technique.

\subsection{Dataset Construction and Model Tested}
Motivated by recent research on the real-world deployment of LLMs in complex downstream reasoning tasks~\cite{tang2024steering, yousuf2024llm}, we construct a benchmark derived from two classical synthetic intelligence analysis challenge datasets: Sign of the Crescent and Atlantic Storm \cite{hughes2003discovery}.
These datasets, originally developed for analyst training, have since become classical benchmarks widely used in analytics competitions and as evaluation datasets in the intelligence analysis and visual analytics research communities \cite{wu2012start, hossain2012storytelling, tahmid2025enhancing, davidson2024investigating}. Each collection offers rich, narrative-style descriptions of individuals, designed to emulate the complexity and ambiguity of real-world intelligence scenarios.

Each data point corresponds to a unique person, represented through a short paragraph containing biographical and contextual details. Entities are implicitly linked through shared activities, affiliations, or co-occurrences, forming the basis of an underlying latent graph.
To establish the ground truth, two annotators manually verified all relational connections between individuals, resulting in a curated graph structure used for evaluation.

We experiment with five models from different model families: i) GPT-4o \cite{hurst2024gpt-4o} from OpenAI, ii) OpenAI o1 \cite{o1_intro, jaech2024openai_o1}, iii) Llama-3 \cite{grattafiori2024llama} from Meta AI, iv) Mistral-7B \cite{jiang2023mistral7b} from MistralAI, and v) Gemini-2 \cite{team2024gemini} from Gemini platform, Google.
\subsection{Can LLMs Recover Structure in Short Contexts?}
\begin{figure*}[t]
\centering
    \begin{subfigure}[t]{0.49\linewidth}
        \centering
        \includegraphics[width=\linewidth]{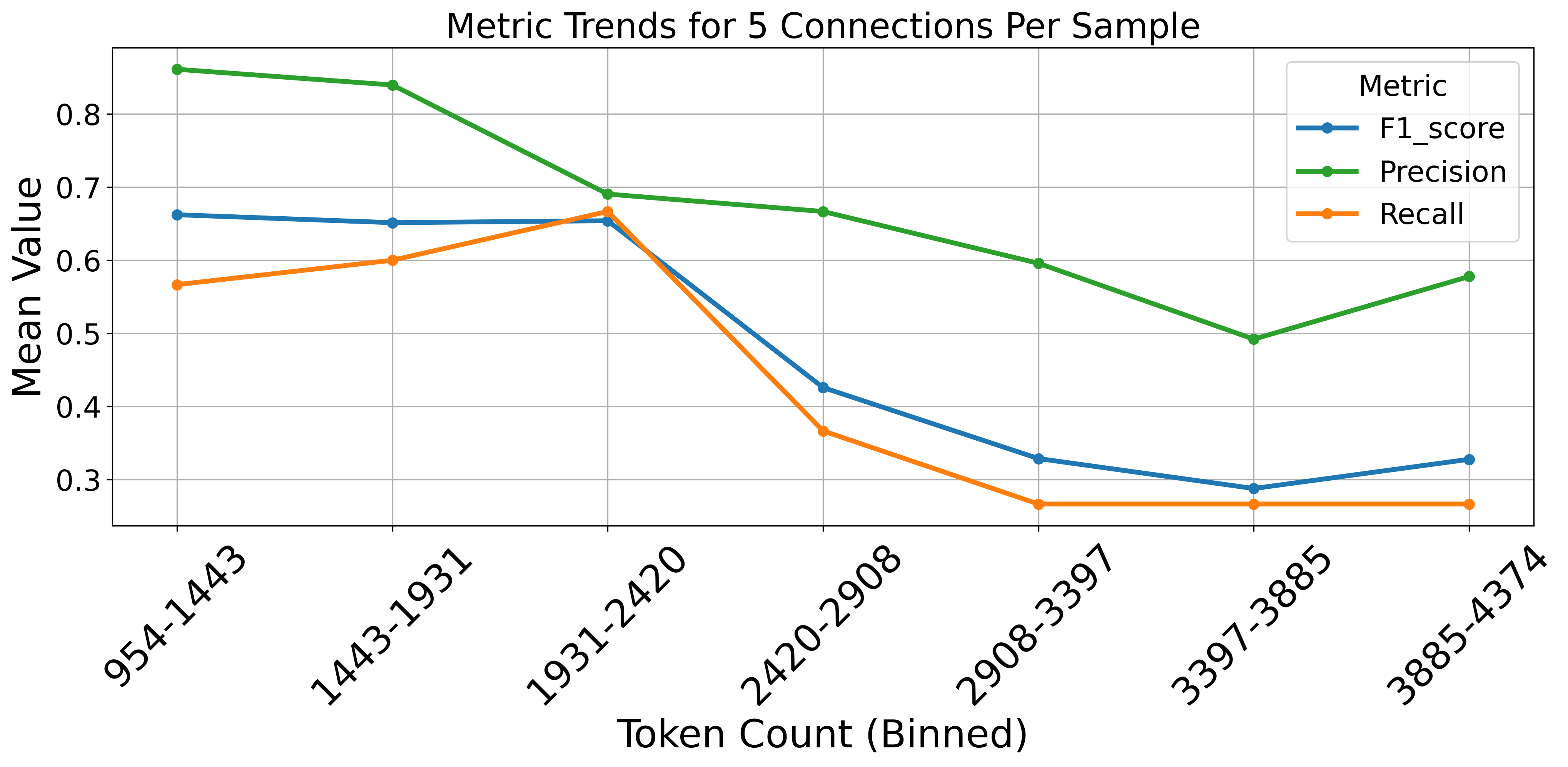}
        \caption{GPT-4o}
        \label{fig:gpt_trend_edge}
    \end{subfigure}
    \hfill
    \begin{subfigure}[t]{0.49\linewidth}
        \centering
        \includegraphics[width=\linewidth]{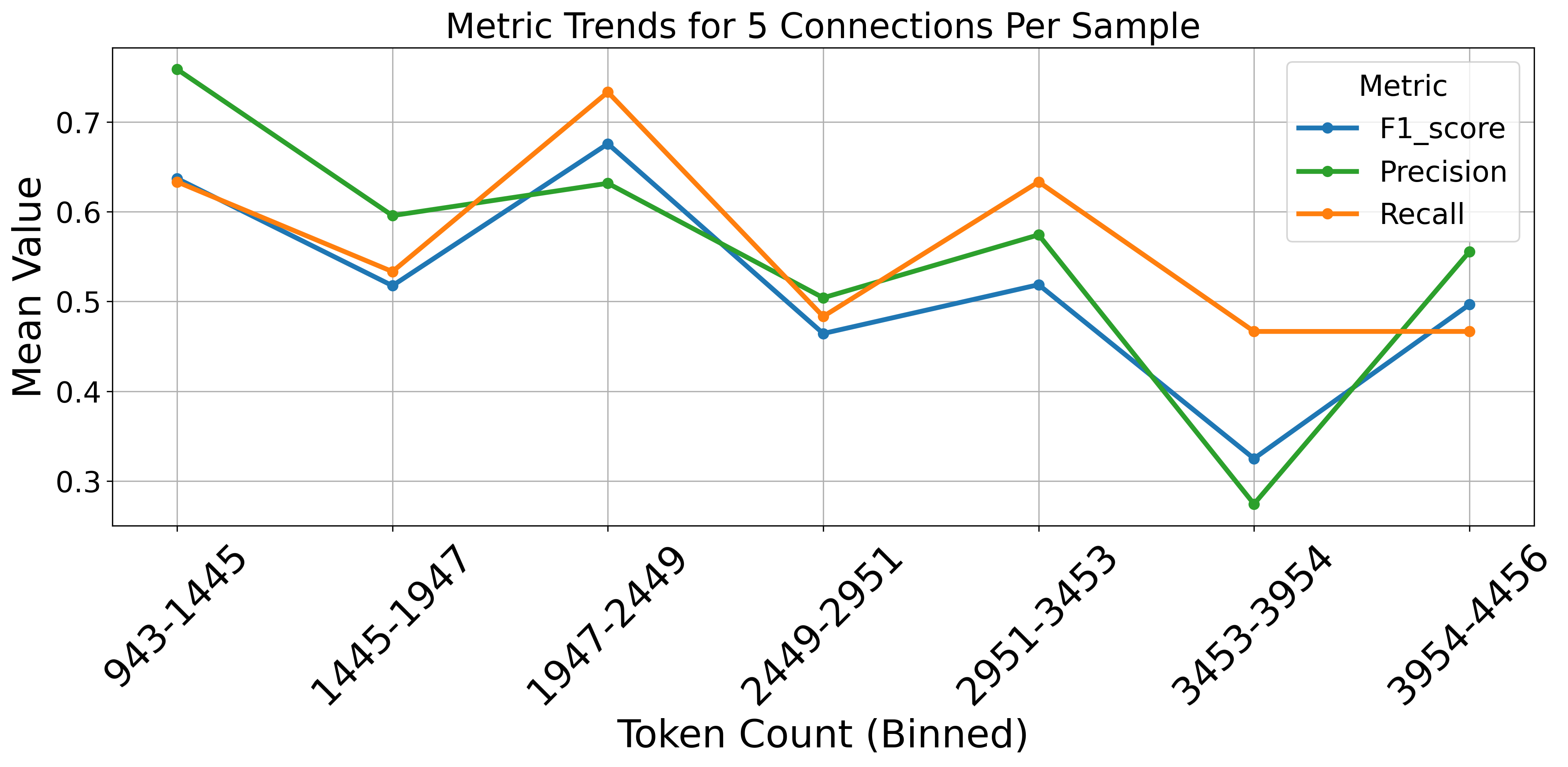}
        \caption{Gemini-2}
        \label{fig:gemini_trend_edge}
    \end{subfigure}
    \caption{Metric trend with GPT-4o and Gemini-2 on edge discovery}
    \label{fig:metric_trend}
\end{figure*}

We begin by evaluating whether LLMs can recover latent graph structure from natural language prompts under idealized conditions, where entities are nearby and context length is short. As shown in Figure~\ref{fig:metric_trend}, GPT-4o and Gemini-2 achieve high precision but only moderate recall and F1. This indicates that the models avoid hallucinating structure, but generaly fail to retrieve many true connections. While partial structural reasoning is evident, complete graph reconstruction remains elusive even in low-memory-stress settings.

We now turn to more challenging conditions, where increased context length, dispersion, and structural density begin to degrade recovery.
\begin{table*}[t]
\caption{Examples of the memory drift metric under different graph reconstruction scenarios. Drift penalizes both hallucinations (FP) and forgetting (FN), offering a more comprehensive signal than standard precision and recall alone.}
\centering
\resizebox{0.95\textwidth}{!}{%
\begin{tabular}{@{}lcccccccc@{}}
\toprule
\textbf{Example} & \textbf{Gold Edges} & \textbf{Predicted Edges} & \textbf{TP} & \textbf{FP} & \textbf{FN} & \textbf{Precision} & \textbf{Recall} & \textbf{Memory Drift} \\
\midrule
\textbf{Perfect (0.00)} & \{(A,B), (B,C)\} & \{(A,B), (B,C)\} & 2 & 0 & 0 & 1.00 & 1.00 & 0.00 \\
\textbf{Mid-case (0.50)} & \{(A,B), (B,C), (C,D)\} & \{(A,B), (C,D)\} & 2 & 0 & 1 & 1.00 & 0.67 & 0.50 \\
\textbf{Balanced (0.75)} & \{(A,B), (B,C), (C,D), (D,E)\} & \{(A,B), (C,D)\} & 2 & 0 & 2 & 1.00 & 0.50 & 0.75 \\
\textbf{Hallucinated (0.875)} & \{(A,B), (B,C)\} & \{(A,B), (A,C)\} & 1 & 1 & 1 & 0.50 & 0.50 & 0.875 \\
\textbf{None (1.00)} & \{(A,B), (B,C)\} & \{\} & 0 & 0 & 2 & 0.00 & 0.00 & 1.00 \\
\bottomrule
\end{tabular}
}
\label{tab:drift_examples}
\end{table*}

\section{Memory Drift and Degradation}
To assess a model’s ability to recover relational structure from noisy input, we report standard edge-level retrieval metrics, \textit{precision}, \textit{recall}, and \textit{F1 score} alongside a primary diagnostic measure we call \textit{memory drift}. While the former reflect local prediction quality, memory drift is designed to capture global performance degradation under increasing context length and relational complexity.

\textbf{Memory Drift} quantifies deviation from ideal relational reconstruction using a weighted sum of true positives (TP), false positives (FP), and false negatives (FN), ignoring true negatives due to their overwhelming presence and low informativeness. The metric is defined as:
\[\scriptsize{
\text{Memory Drift} = 1 - \max\!\left(0,\ \frac{w_{\text{TP}} \cdot \text{TP} \;+\; w_{\text{FP}} \cdot \text{FP} \;+\; w_{\text{FN}} \cdot \text{FN}}{2P} \right)}\]
where $P$ is the number of gold-standard edges in the prompt. The weights $w_{\text{TP}} = 2$, $w_{\text{FP}} = -0.5$, and $w_{\text{FN}} = -1.0$ reflect our view that forgetting (missed edges) is more damaging than hallucination (spurious edges), though both degrade structural integrity. The $\max(0, \cdot)$ clamp ensures that large numbers of errors do not yield negative values. This formulation produces a bounded value in $[0, 1]$, where $0$ indicates perfect structural recovery and $1$ indicates maximal degradation. 
To build intuition for how memory drift behaves in practice, Table~\ref{tab:drift_examples} shows example predictions with varying combinations of true positives, false positives, and false negatives. These examples illustrate how memory drift increases as models forget edges, hallucinate new ones, or both, even when standard precision and recall appear reasonable.

Importantly, \textit{memory drift is not equivalent to recall}. A model may exhibit reasonable recall but still show high drift if it frequently introduces incorrect structure. By incorporating both types of errors, the metric captures a broader notion of degradation relevant to downstream reasoning tasks.

To support interpretability, we also report standard metrics:

\[
\begin{array}{@{}l@{\hspace{2cm}}r@{}}
\text{Precision} = \dfrac{\text{TP}}{\text{TP} + \text{FP}} &
\text{Recall} = \dfrac{\text{TP}}{\text{TP} + \text{FN}}
\end{array}
\]
\[
\text{F1} = \frac{2 \cdot \text{Precision} \cdot \text{Recall}}{\text{Precision} + \text{Recall}}
\]

These standard retrieval metrics allow us to disentangle false positives and false negatives. All four measures are tracked across token lengths and connection densities to characterize long-context relational reasoning.
\begin{figure*}[t]
\centering
    \begin{subfigure}[t]{0.32\linewidth}
        \centering
        \includegraphics[width=\linewidth]{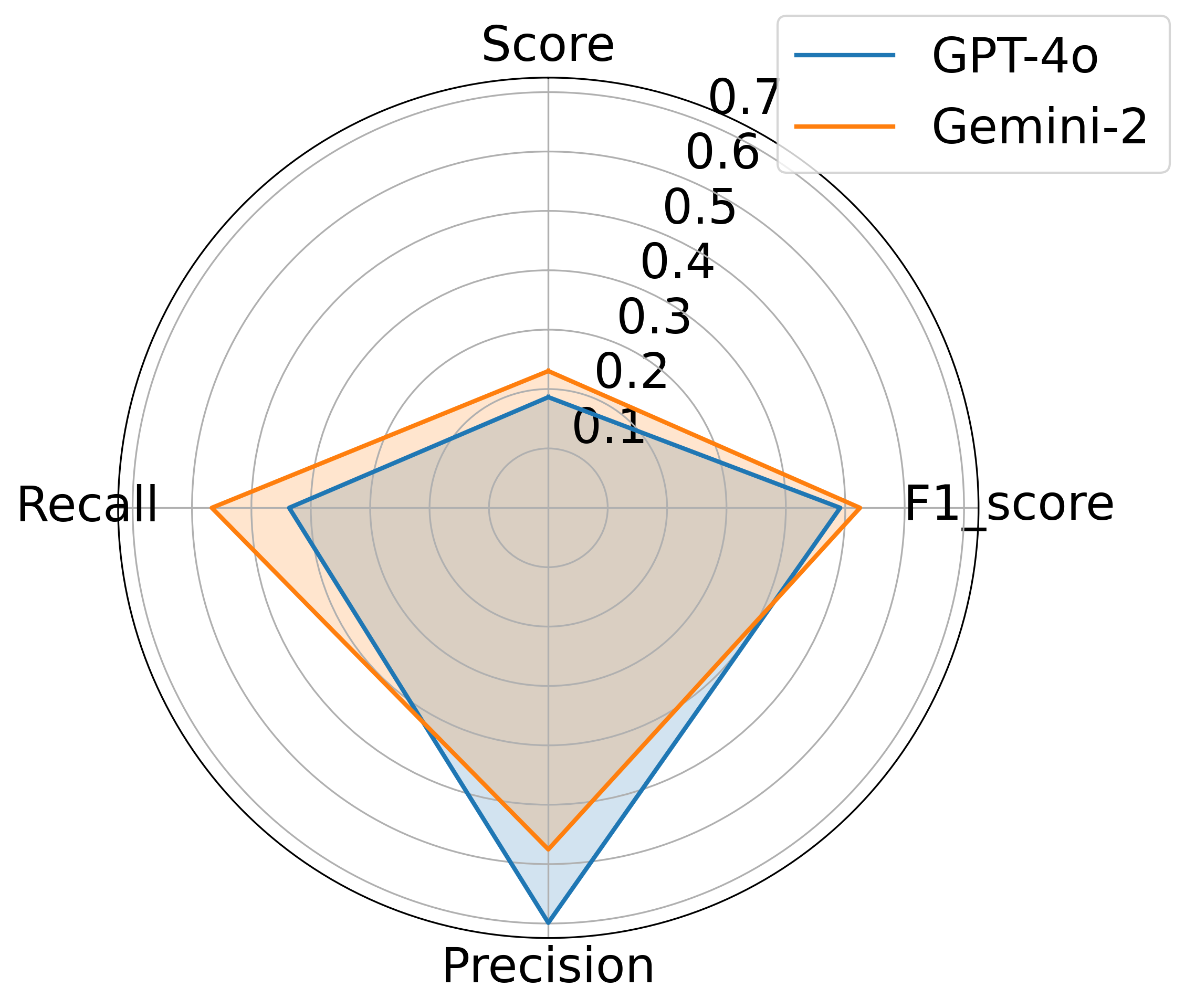}
        \caption{GPT-4o and Gemini-2 (edge)}
        \label{fig:gpt_radar_edge}
    \end{subfigure}
    \hfill
    \begin{subfigure}[t]{0.32\linewidth}
        \centering
        \includegraphics[width=\linewidth]{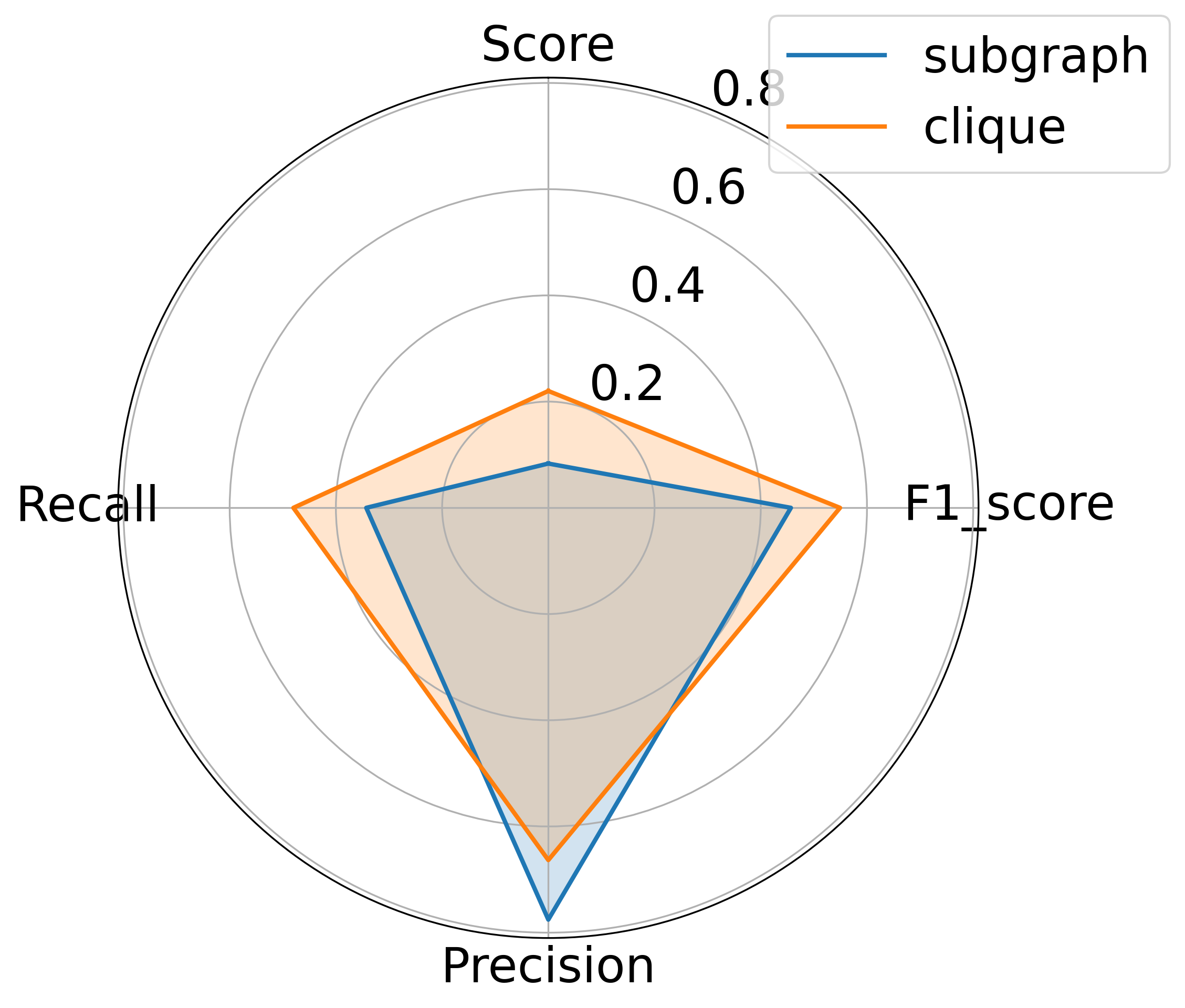}
        \caption{GPT-4o(subgraph and clique)}
        \label{fig:gpt_overlay}
    \end{subfigure}
    \hfill
    \begin{subfigure}[t]{0.32\linewidth}
        \centering
        \includegraphics[width=\linewidth]{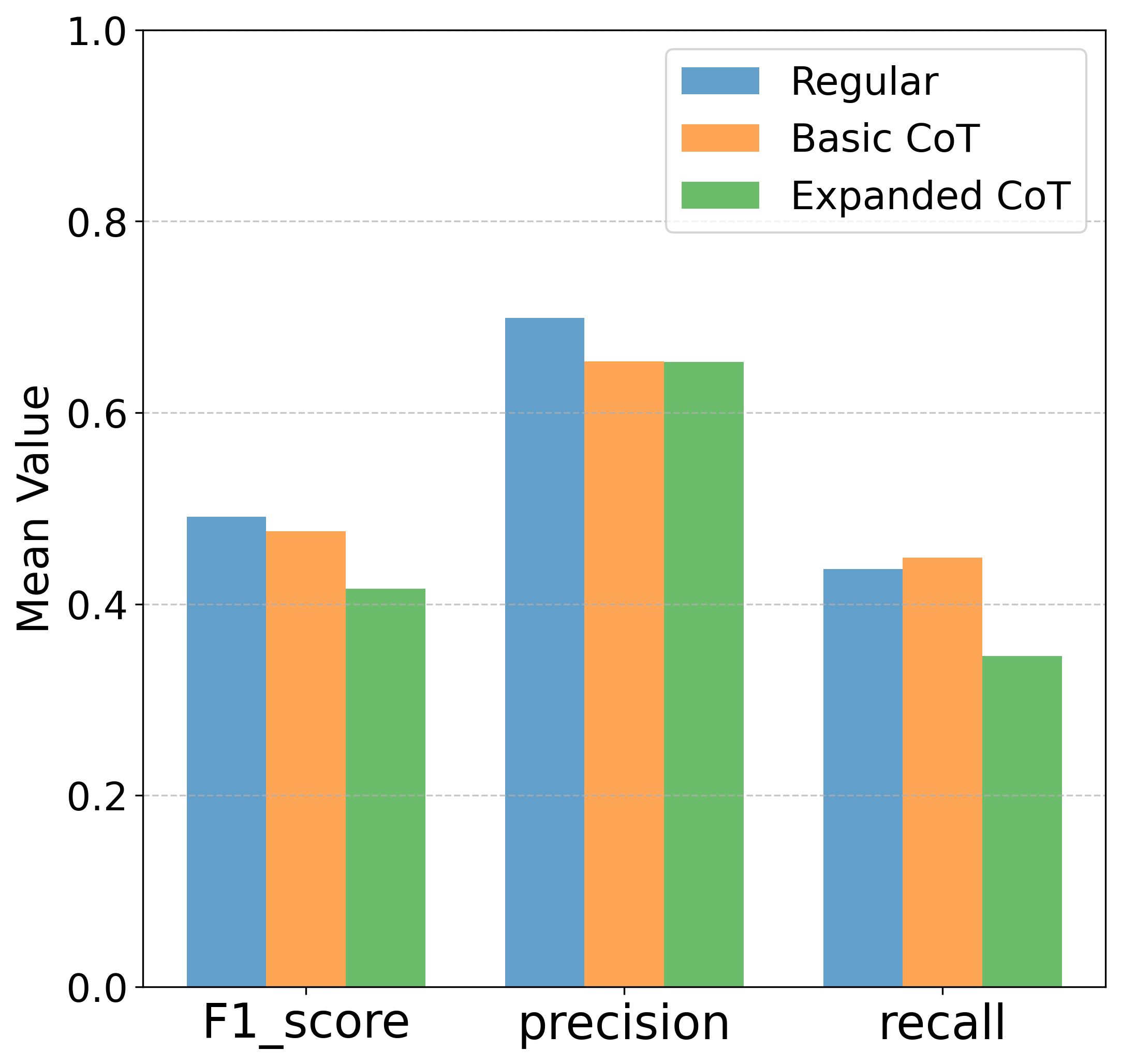}
        \caption{Different prompting style}
    \label{fig:gpt_cot_bar_edge}
    \end{subfigure}
    \caption{Metrics for different model and discovery cases, ``Score'' corresponds to \((1 - \text{Memory Drift})\), allowing comparison with precision, recall, and F1.}
    \label{fig:radar}
\end{figure*}

\begin{figure*}[t]
\centering
    \begin{subfigure}[t]{0.49\linewidth}
        \centering
        \includegraphics[width=\linewidth]{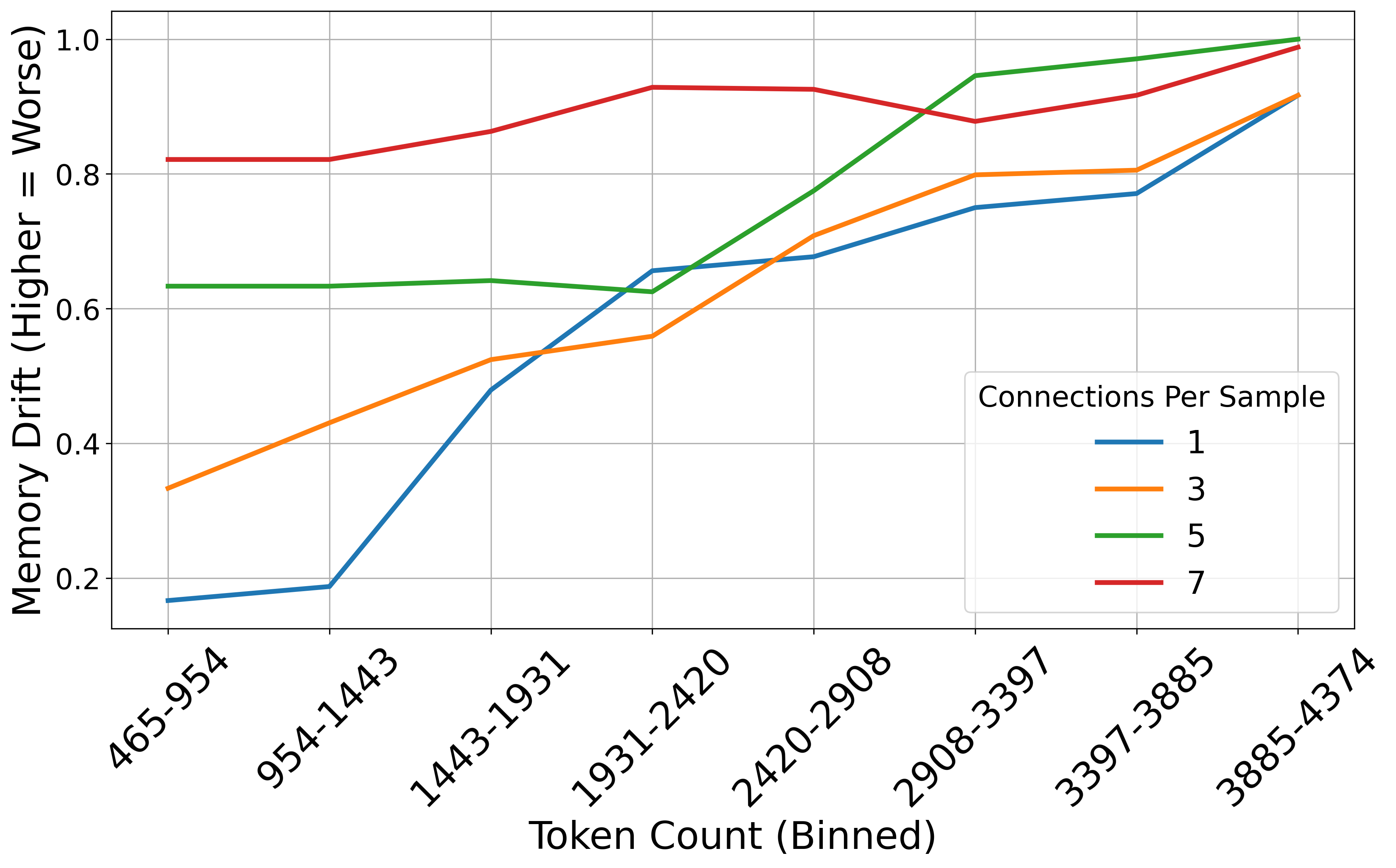}
        \caption{Memory drift for GPT-4o}
        \label{fig:gpt_drift_edge}
    \end{subfigure}
    \hfill
    \begin{subfigure}[t]{0.49\linewidth}
        \centering
        \includegraphics[width=\linewidth]{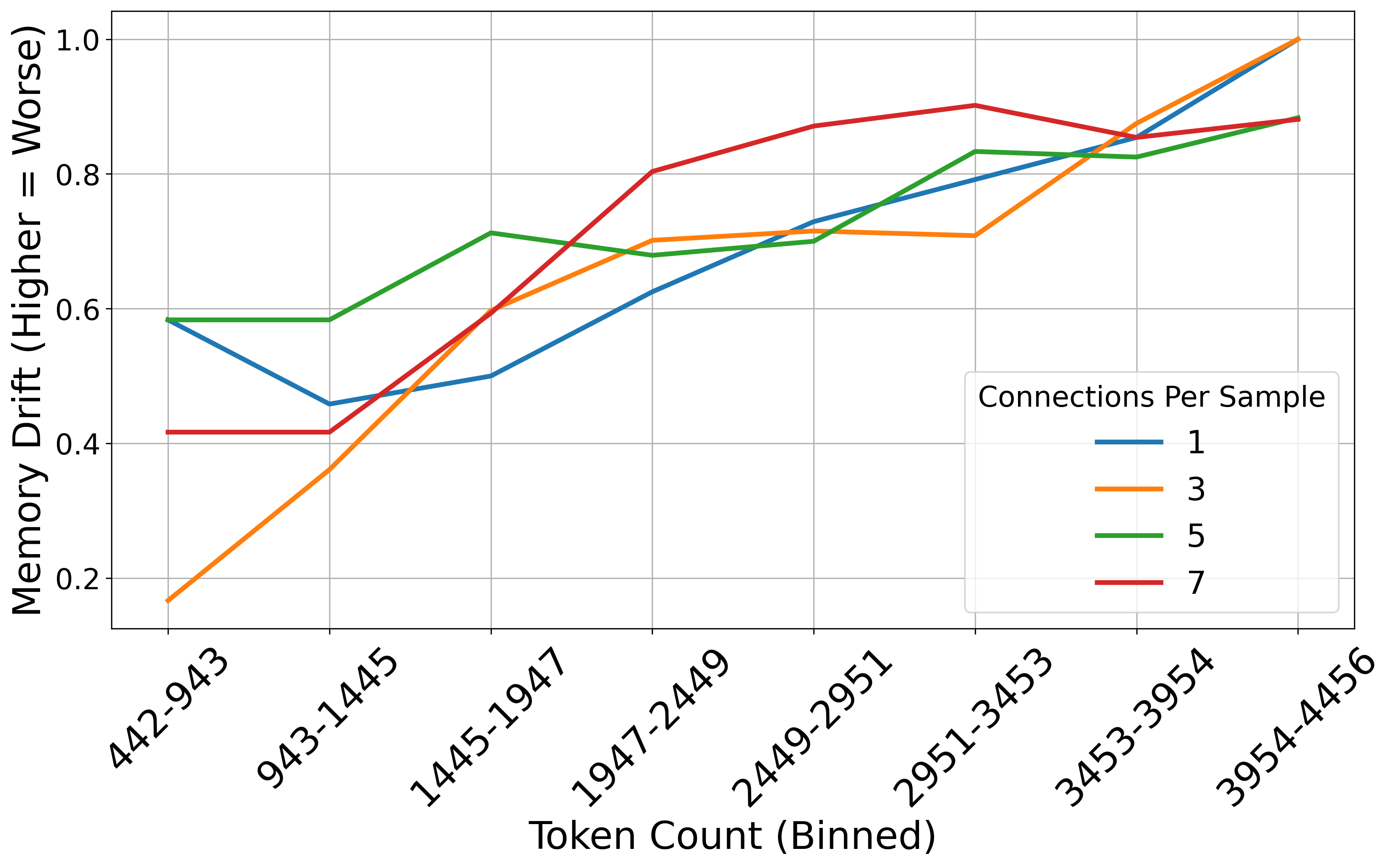}
        \caption{Memory drift for Gemini-2}
        \label{fig:gemini_drift_edge}
    \end{subfigure}
    \caption{Memory drift on edge discovery for different models}
    \label{fig:mem_drift}
\end{figure*}

\subsection{When Does Memory Drift Begin?}
We analyze how structural recovery degrades with increasing input length. Memory drift refers to this performance decay as related entities become more contextually separated.

As shown in Figure~\ref{fig:gpt_drift_edge} for GPT-4o and Figure \ref{fig:gemini_drift_edge} for Gemini-2, \textbf{memory drift increases sharply after 2000 tokens across all connection densities}. Low-density prompts suffer the fastest degradation, while higher-density ones appear more stable but begin with worse initial performance.

Figure~\ref{fig:gpt_radar_edge} for GPT-4o shows that degradation is driven by declining recall, while precision remains stable. This suggests that models increasingly miss true edges rather than hallucinating new ones. The onset of memory drift at around 2000 tokens marks a practical upper bound on effective context for relational reasoning. However, Gemini-2 shows a more balanced behavior but still suffers high memory drift after 2000 tokens.

\subsection{Does Information Density Help or Hurt?}
\label{ssec:info_density}
\textbf{Information density, measured by the number of connections per sample, negatively affects model performance.} As density increases, the task becomes more difficult due to greater relational complexity.

Both Figure~\ref{fig:mem_drift} and Figure~\ref{fig:metric_trend} show that high-density prompts begin with lower F1 score and higher memory drift, even at short context lengths. This indicates that models struggles to recover dense structures regardless of token count. The effect compounds over longer contexts, but the primary impact is already visible in the initial token bins. The degradation for higher density prompts is more prominent in GPT-4o. Gemini-2 shows a more stable behavior for higher density prompts.

\subsection{Hallucination vs Forgetting}
We assess how models balance false positives and false negatives under increasing context length and connection density. Specifically, we examine whether performance degradation is driven by hallucinated edges (low precision) or missed ones (low recall).

As shown in Figure~\ref{fig:gpt_trend_edge}, GPT-4o maintains consistently high precision across all token bins, even as recall decline sharply with longer contexts. This pattern holds across densities and is further supported by the radar plot in Figure~\ref{fig:gpt_radar_edge}, where precision dominates all other metrics.

These results suggest that the \textbf{model adopts a conservative prediction strategy}. It prefers to omit uncertain connections rather than risk false positives. Hallucination remains rare, even in dense or noisy prompts, and does not increase with token count. Instead, most errors arise from failure to recover true edges, particularly under dispersion and structural complexity.

In summary, hallucination is not the dominant failure mode in long-context relational reasoning. The model prioritizes precision at the expense of recall, leading to underprediction rather than overgeneration.
\begin{figure*}[t]
\centering
    \begin{subfigure}[t]{0.49\linewidth}
        \centering
        \includegraphics[width=\linewidth]{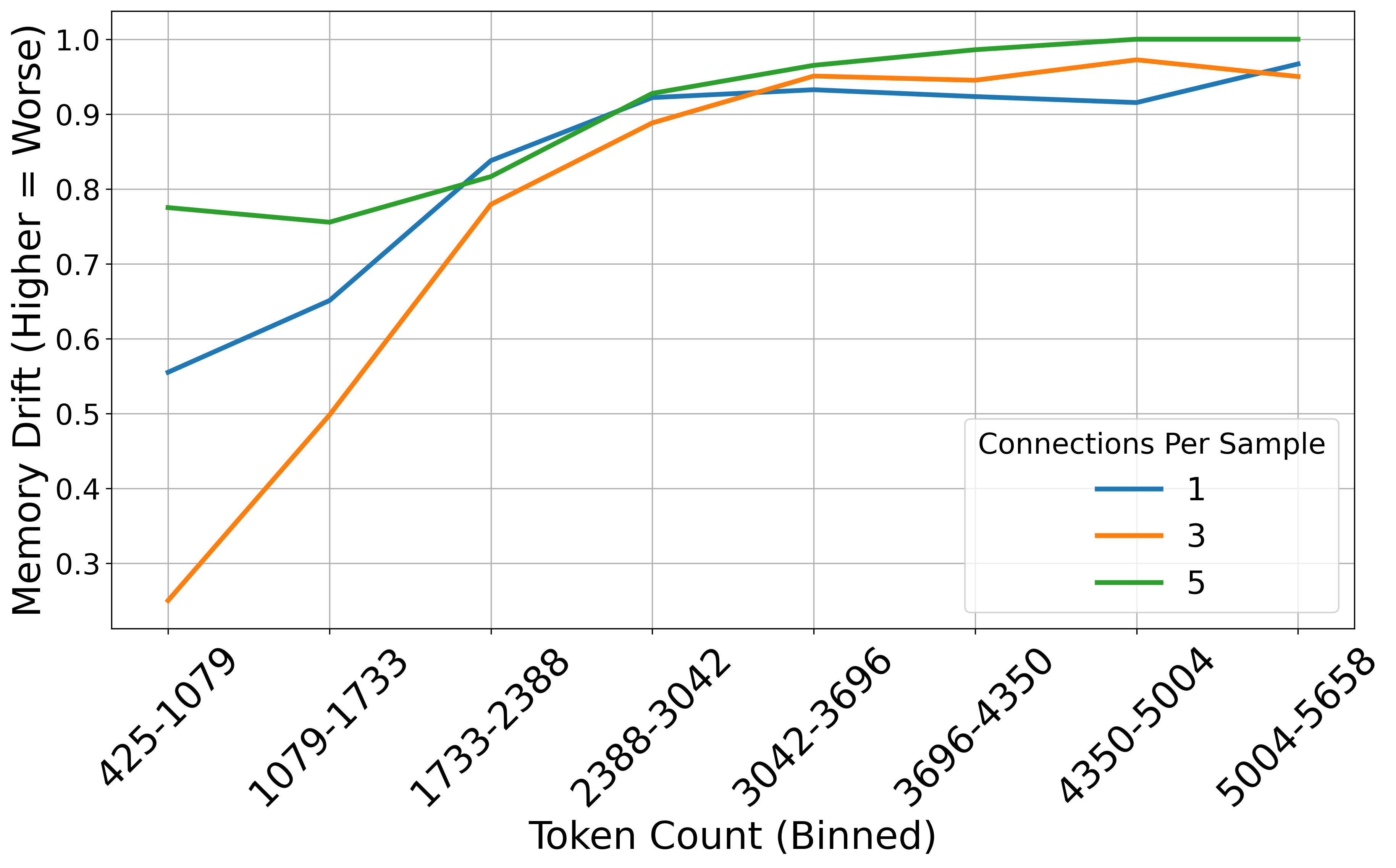}
        \caption{Memory drift on subgraph discovery}
        \label{fig:gpt_drift_subgraph}
    \end{subfigure}
    \hfill
    \begin{subfigure}[t]{0.49\linewidth}
        \centering
        \includegraphics[width=\linewidth]{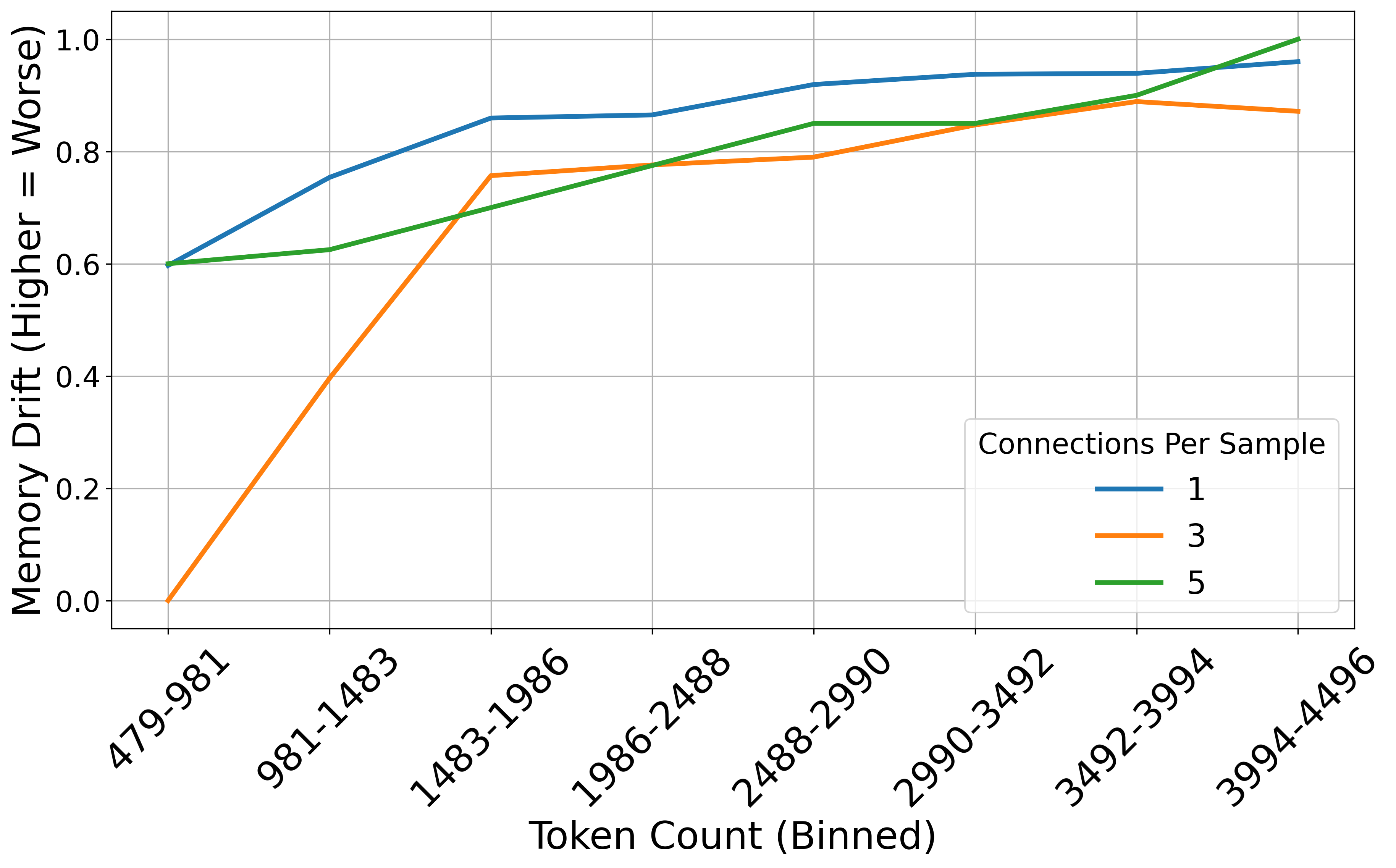}
        \caption{Memory drift on clique discovery}
        \label{fig:gpt_drift_clique}
    \end{subfigure}
    \caption{Memory drift for GPT-4o on two different subtasks}
\end{figure*}

\subsection{Does Chain-of-thought (CoT) Prompting Help?}
We test whether Chain-of-Thought (CoT) \cite{wei2022chain} prompting improves structural extraction by comparing regular prompting, basic CoT, and expanded CoT on the edge discovery task. As shown in Figure~\ref{fig:gpt_cot_bar_edge} for GPT-4o, both CoT variants underperform the regular strategy across all key metrics. Expanded CoT performs the worst, with notably lower score, recall, and F1. Basic CoT shows slightly better recall than regular, but at the cost of reduced overall accuracy.

We conclude that \textbf{CoT prompting is not helpful for this task}. For relational recovery in long contexts, simple prompting remains the most effective approach.

\begin{table*}
\caption{Model comparison across long-context relational reasoning benchmarks. }
\centering
\resizebox{\textwidth}{!}{%
\begin{tabular}{@{}lcccc@{}}
\toprule
\textbf{Capability} & \textbf{GPT-4o} & \textbf{Gemini-2} & \textbf{Llama-3} & \textbf{Mistral-7B} \\
\midrule
\textbf{Memory Drift Onset} & $\sim$2000 tokens & $\sim$2500 tokens & $\sim$1800 tokens & $<$1000 tokens \\
\textbf{Precision} & \textcolor{green}{Very High (stable)} & \textcolor{orange}{Moderate (fluctuating)} & \textcolor{green}{High (slightly unstable)} & \textcolor{red}{Low (inconsistent)} \\
\textbf{Recall} & \textcolor{red}{Low (declines early)} & \textcolor{green}{High (adaptive)} & \textcolor{orange}{Moderate (variable)} & \textcolor{red}{Low (noisy)} \\
\textbf{F1 Score Stability} & \textcolor{orange}{Moderate (recall-limited)} & \textcolor{green}{Balanced (peaks mid-context)} & \textcolor{orange}{Moderate (flat)} & \textcolor{red}{Unstable} \\
\textbf{Density Robustness} & Poor beyond 5 connections & Robust up to 7 connections & Moderate (degrades after 5+) & Fails even at 3+ connections \\
\textbf{Prediction Strategy} & Conservative (high precision) & Balanced (recall-oriented) & Slightly conservative & Inconsistent \\
\textbf{Best Use Case} & Precision-critical tasks & Broad structure recovery & General-purpose tasks & Lightweight/fine-tuned use \\
\bottomrule
\end{tabular}
}
\label{tab:model_comparison}
\end{table*}

\subsection{LLMs' Behavior across Different Graph Discovery Subtasks}
We compare GPT-4o's performance across the three relational reasoning subtasks: edge discovery, subgraph (degree-based) discovery, and clique discovery. These differ in structural complexity and the type of inductive reasoning required.

Figure~\ref{fig:gpt_radar_edge} and \ref{fig:gpt_overlay} show metric profiles for each task under the same connection density (5 connections per sample). Edge recovery yields the highest precision but lowest recall, suggesting that the model avoids hallucination but misses many valid edges. Degree-based subgraph discovery shows more balanced recall and precision, though overall scores remain low. Clique recovery exhibits the highest recall and F1, but struggles with precision due to the combinatorial challenge of predicting fully connected structures.

Memory drift curves (Figure~\ref{fig:gpt_drift_edge}, \ref{fig:gpt_drift_subgraph}, and \ref{fig:gpt_drift_clique}) reveal that drift patterns vary by task. Edge and degree-based subtasks degrade quickly beyond 2000–2500 tokens, while clique recovery declines more slowly, likely due to redundancy among densely connected entities. These findings highlight that task structure strongly affects how LLMs handle long-context reasoning. Sparse graphs lead to brittle recall, dense graphs trigger hallucination risk, and mid-level structures offer moderate balance but still degrade under dispersion.

\subsection{Recommendations for Different LLMs}
Our evaluation reveals that different models exhibit distinct tradeoffs in how they handle long-context relational reasoning. Table~\ref{tab:model_comparison} summarizes these trends across key behavioral axes, including memory drift onset, precision-recall balance, and robustness to relational density. 
While no model is uniformly superior, each demonstrates strengths in specific regimes. Further figures and details are available in our codebase. Based on these observations, we offer the following practical recommendations for model selection, tailored to the needs of different real-world use cases:
\begin{itemize}
    \item \textbf{GPT-4o} is well-suited for high-precision tasks where hallucination must be minimized, such as legal, intelligence vetting, or safety critical settings. However, it sacrifices recall as context length and complexity grow.
    \item \textbf{Gemini-2} is ideal for exploratory tasks that prioritize coverage, such as the construction of a knowledge graph or intelligence discovery. It shows high recall and better robustness under structural noise.
    \item \textbf{Llama-3} offers strong precision with moderate recall, making it suitable for general-purpose long-context reasoning. It handles moderate relational density well but begins to degrade with structural complexity and extended dispersion.
    \item \textbf{Mistral-7B} struggles in zero-shot relational recovery but may serve as a lightweight base for fine-tuning or retrieval-augmented pipelines, particularly when resources are constrained.
\end{itemize}

\begin{figure}[t]
    \centering
    \includegraphics[width=0.70\linewidth]{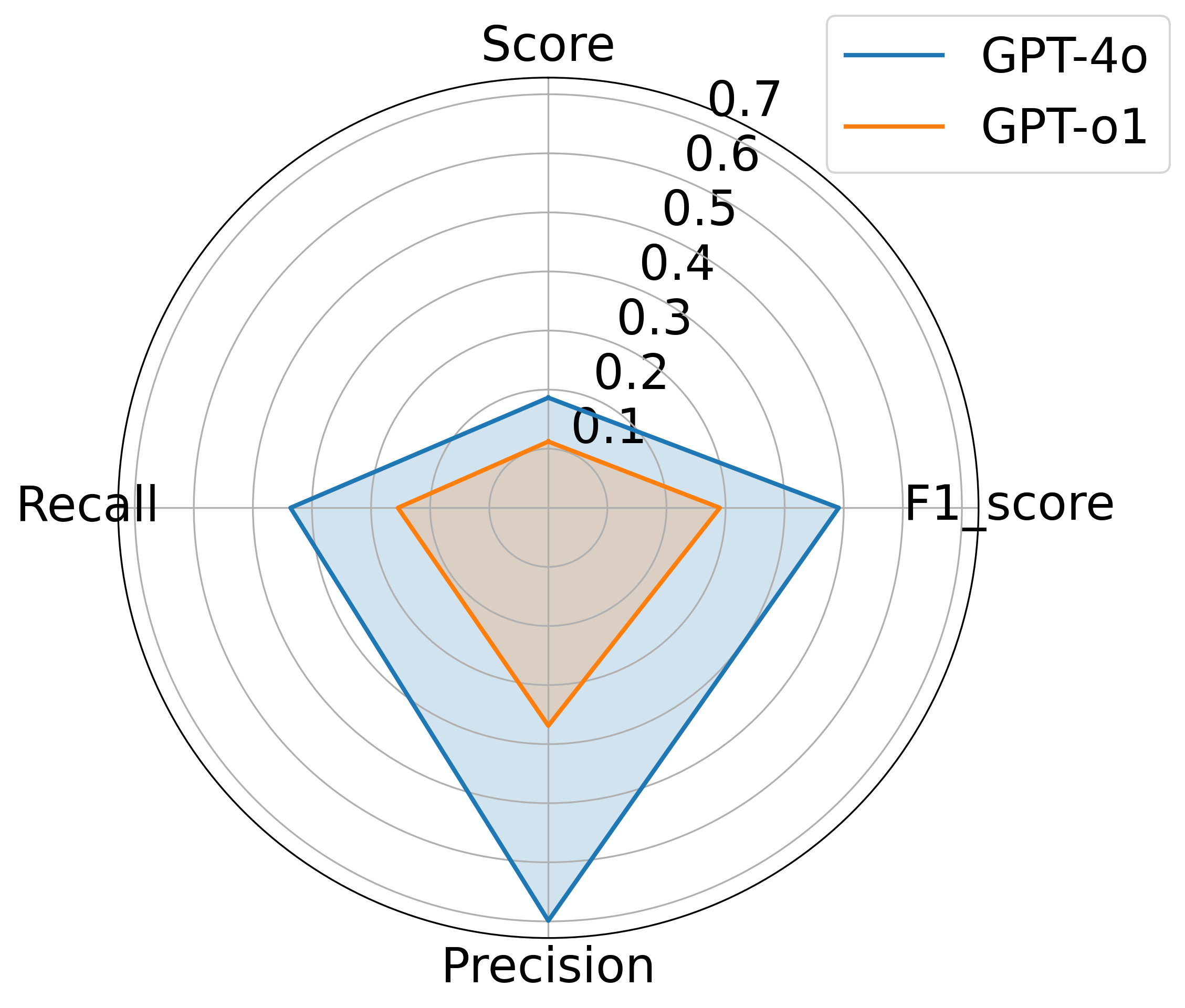}
    \caption{Metrics for GPT-4o and o1 (edge) cases, ``Score'' corresponds to \((1 - \text{Memory Drift})\), allowing comparison with precision, recall, and F1.}
    \label{fig:gpto1_radar_edge}
\end{figure}

\begin{figure}[t]
    \centering
    \includegraphics[width=\linewidth]{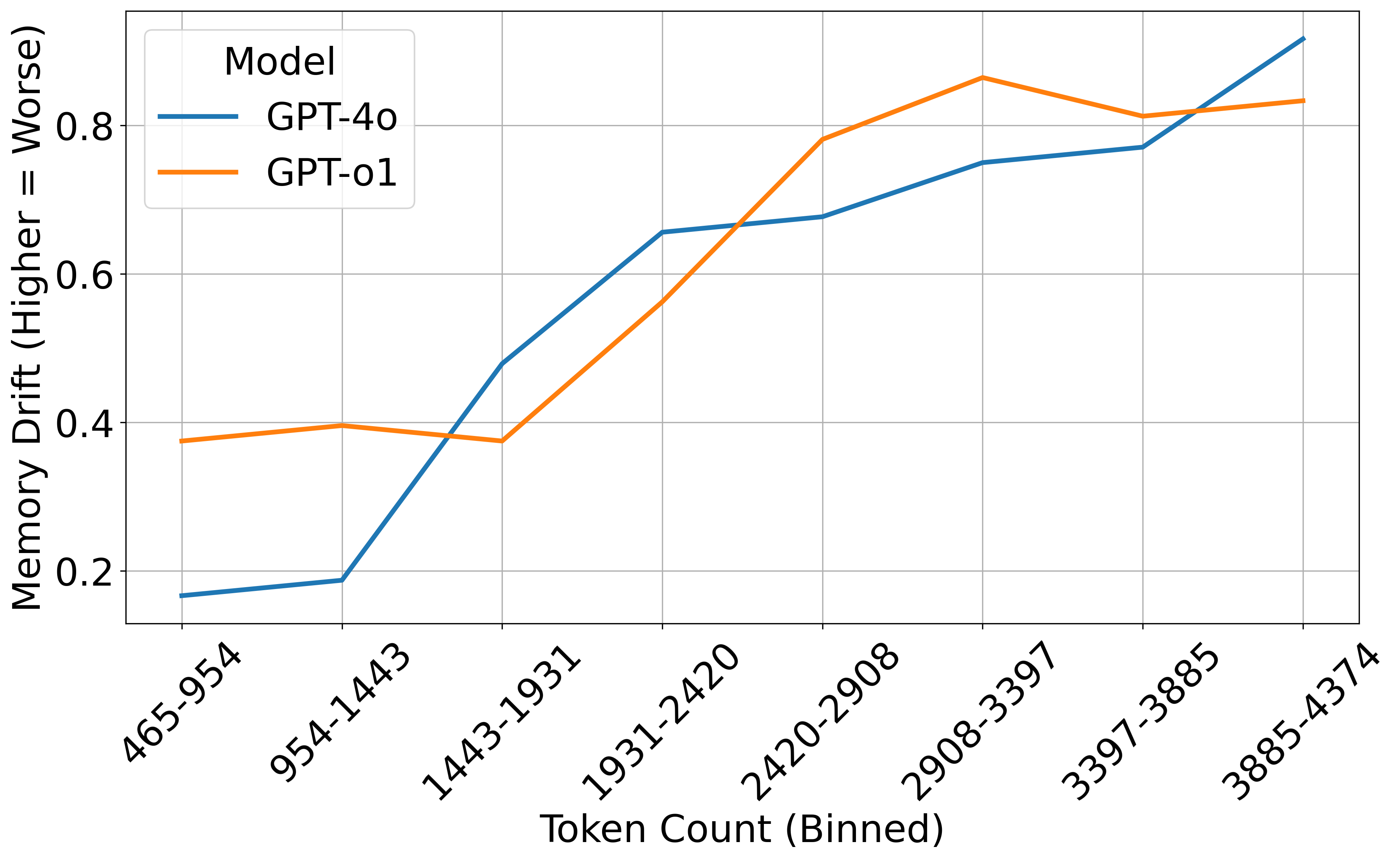}
    \caption{Memory drift GPT-4o and o1 (edge) case}
    \label{fig:gpto1_memory_drift}
\end{figure}

\subsection{Can Reasoning-Oriented Models Like OpenAI o1 Overcome Memory Drift?}
\textbf{OpenAI o1 does not outperform general-purpose models on memory drift, and suffers similar or worse degradation as context length increases.}

We assess whether reasoning-specialized models, such as o1 \cite{o1_intro, jaech2024openai_o1}, are more robust to memory drift on long-context relational reasoning tasks.

As shown in Figure~\ref{fig:entry_overlay}, which presents memory drift for all models, o1 performs comparably to GPT-4o at short context lengths. However, as the token count increases, o1's memory drift rises steadily, matching or exceeding that of the other models. Notably, for the longest contexts, o1 does not outperform general-purpose models and even displays higher drift than GPT-4o and Llama-3 in several bins.

A direct comparison in Figure~\ref{fig:gpto1_memory_drift} (o1 vs. GPT-4o) shows that o1 remains similar to GPT-4o for shorter inputs but its memory drift surpasses GPT-4o for prompts longer than 2000 tokens, confirming that the onset of degradation is not delayed for reasoning-tuned models.

The radar plot in Figure~\ref{fig:gpto1_radar_edge} further illustrates this gap. o1 maintains reasonable precision but lags behind GPT-4o in both recall and F1, indicating that while it avoids hallucination, it fails to recover a significant portion of the true relational structure—especially as input length and complexity increase.

Overall, reasoning-oriented models like o1 do not overcome the limitations of memory drift or context fragmentation in long, noisy inputs. The results indicate that current advances in model reasoning are insufficient for reliable relational graph induction at scale.

\section{Conclusion, Limitations, and Future Work}
We introduced a benchmark for evaluating long-context reasoning in LLMs through the task of graph reconstruction from noisy text. Our results show that models degrade much earlier than their context limits suggest, especially under structural complexity and dispersion. The proposed memory drift metric offers a more accurate view of this degradation than standard retrieval metrics. Our benchmark reveals key tradeoffs across model families and provides a practical lens for assessing real-world reasoning capabilities. These findings offer concrete guidance for both model development and deployment in structure-sensitive applications. 

Several limitations should be acknowledged. Our evaluation is limited to zero-shot and few-shot prompting, without exploring the effects of fine-tuning or retrieval-based approaches. We also recognize that prompt sensitivity is an important consideration and leave a more systematic study of this aspect to future work.

Looking ahead, we will investigate how retrieval-based and memory-augmented systems influence memory retention, forgetting, and drift in long-context relational reasoning. More broadly, we hope this benchmark and metric provide a valuable diagnostic for the nuanced failure modes of current LLMs, moving beyond ``needle in a haystack'' evaluations toward more realistic, structure-based reasoning tasks. By establishing new evaluation settings and highlighting model-specific behaviors, we aim for this work to support continued advances in long-context understanding and structured reasoning.

\bibliographystyle{IEEEtran}
\bibliography{IEEEfull}

\end{document}